# Hand Gesture Recognition for Contactless Device Control in Operating Rooms

Ebrahim Nasr-Esfahani[1], Nader Karimi[1], S.M. Reza Soroushmehr[2,3],
M. Hossein Jafari[1], M. Amin Khorsandi[1], Shadrokh Samavi[1,3], Kayvan Najarian[2,3,4]

[1]Department of Electrical and Computer Engineering, Isfahan University of Technology, Isfahan 84156-83111, Iran
[2]Michigan Center for Integrative Research in Critical Care, University of Michigan, Ann Arbor, 48109 U.S.A
[3]Department of Emergency Medicine, University of Michigan, Ann Arbor, U.S.A
[4]Department of Computational Medicine and Bioinformatics, University of Michigan, Ann Arbor, 48109 U.S.A

**Abstract.** Hand gesture is one of the most important means of touchless communication between human and machines. There is a great interest for commanding electronic equipment in surgery rooms by hand gesture for reducing the time of surgery and the potential for infection. There are challenges in implementation of a hand gesture recognition system. It has to fulfill requirements such as high accuracy and fast response. In this paper we introduce a system of hand gesture recognition based on a deep learning approach. Deep learning is known as an accurate detection model, but its high complexity prevents it from being fabricated as an embedded system. To cope with this problem, we applied some changes in the structure of our work to achieve low complexity. As a result, the proposed method could be implemented on a naive embedded system. Our experiments show that the proposed system results in higher accuracy while having less complexity in comparison with the existing comparable methods.

## 1 Introduction

Communication with an electronic device without using hand is an open problem which has lots of advantages. Convenience and maintenance of hygiene standards are the most important advantages of such a system. This field has become of interest specifically in the surgery rooms to help surgeons while keeping their hands sterile. An example is "Gestix", a medical instrument consisting of a large display for browsing MRI image. It receives its commands from a human operator through hand gestures [1]. The most important advantage of this system is keeping surgeons and staff's hand sterile during the surgery. Another example of hand gesture application in surgery rooms is proposed in [2]. Authors in [2] implement a system of hand gesture recognition to command a robot called "Gestonurse". This robot is in charge of delivering surgical instruments to surgeons during the surgery. Experiments done in [3] indicate that hand gesture commands are faster and more accurate than voice commands to a scrub nurse. Another work in [3] proposes a system of controlling lights in the operation room. Feasibility of working with a hand tracking system and commanding the device without touching non-sterile spots are the most important advantages of this system.

Communication of surgeons or hospital staff with these devices without any keyboard or mouse (which are important means of spreading infections) maintains standards of hygiene in hospital and surgery rooms as well. Voice command is another type of touchless communication but its commands are discrete rather than hand gestures which are able to perform analog commands [4]. On the other hand, voice command has other disadvantages such as its low accuracy due to existence of noise in surgery rooms [5]. Although it seems that these devices fulfill the desires of people completely, their functionality is restricted due to their limited number of supported gestures [4]. To cope with this problem, some alternatives are proposed such as two-hand gesture or combined hand gestures with voice commands. An example of what a hand gesture system could consist of is shown in Fig. 1, where the surgeon could command a display to show different X-ray images.

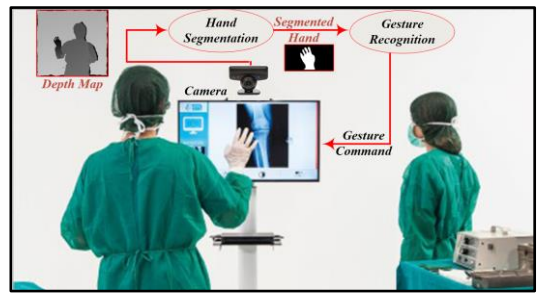

Fig. 1. Image depth is used for recognition of surgeon's hand gestures. Background image is from [6].

Hand gesture recognition systems usually consist of 3 phases of detection, tracking, and recognition [7]. The detection phase is responsible for segmentation of the hand region. Well known inputs for this problem are depth and color. The role of tracking phase is to produce a model of hand movement by providing a frame-to-frame model of hand or its features. The tracking phase is also used for estimation of features and parameters which are not available in a high speed system. The final and the most important phase in hand gesture recognition is its recognition phase. The goal of this phase is to interpret the message in which the gesture is trying to convey. The recognition phase is an open field of study and many researches with different definitions are proposed for this issue. The work in [8] performs a boundary analysis algorithm to locate convex points. These convex points are exploited to detect finger tips and finger valleys for final recognition. The work in [9] uses machine learning technique to its recognition phase. Bag-of-features and support vector machine (SVM) are key structures of recognition in this system. Another work in [10] is mainly focused on finger detection considering geometry of fingers. The fingers are modeled as a cylindrical shape with parallel lines and both open and closed fingers are detected. Proposed work in [11] acquires data of hand from 3 devices including Kinect, camera, and a leap motion sensor. The collected data are contributed to produce some geometric features of hand. These features consist of finger features (angle, distance, elevation and fingertip), hand palm center, hand orientation, histogram of distance, and the curvature of hand contour. The extracted features are fed to an SVM for classification and final recognition. Complexity and the need to use several devices are drawbacks of this method. A few works are proposed to cope with the problem of hand gesture recognition by the use of hardware. The work in [12] proposed an architecture for implementing hand gesture recognition on an FPGA. This work concentrated on detection of finger tips by considering hand geometries. The output of this structure is the open fingers count which is not sufficient for a hand gesture recognition system.

One of the most important goals of researches in the field of hand gesture is to achieve real-time responses. But real-time response requires fast algorithms which sometimes suffer from low accuracy. However, implementation of a real-time hand gesture recognition system with accurate response is desirable in a place like surgery room. One important requirement of real-time systems is simplicity for implementation in an embedded system. Non-real-time operational systems are not trustworthy for vital functions due to their multi-task structures and possibility of halt in their operation [13]. Embedded systems are usually used for systems which run only one task. They should be reliable and fast. Although, design of an embedded system is harder than implementation of an algorithm on a computer, but advantages of embedded systems outweigh their difficult design process. The design of an embedded system can be based on software, using processors and microcontrollers, or it can be based on hardware, using FPGA and ASIC devices. Hardware implementation of image processing algorithms has many benefits as compared to software implementation [14]. These advantages consist of less power consumption, faster functionality, and higher reliability in noisy environments. Authors of [14] put an emphasis on the remarkable aspects of image processing on hardware and the use of convolution filters for hardware implementation. They show that such hardware implementation has faster and less computational complexity than the software version.

Deep learning methods are widely used for solving different aspects of pattern recognition problems [15, 16]. Convolutional neural networks (CNNs) are among most powerful deep learning approaches applicable in different imaging fields. CNNs can be used as an effective method for automatic extraction of discriminative features from large amount of sample images. Recently CNNs have shown significant improvements in different medical imaging applications [17-19] and are specifically effective in recognition and differentiation of various structures. This was the main reason that we considered to implement such a network for recognizing different hand gestures.

In this paper, we propose a method based on deep neural networks to tackle the hand gesture recognition problem for surgical room applications. In the first step in our proposed system, the hand region is detected and extracted as a binary mask. The segmentation is performed based on depth information provided by the camera. The binary map, showing structure of the hand, is applied to a CNN. A method for augmentation of available data is used for enhancing the CNN's performance. The CNN classifies the images of gestures into 10 defined classes. Moreover, a method for binarizing the network is applied. The binarization would simplify the network for implementation as a part an embedded hardware system. Experimental results demonstrate that our proposed system outperforms existing comparable methods in terms of recognition accuracy and implementation complexity.

İn the remainder of this paper, in Section 2 the proposed method for recognition of hand gestures is explained. Section 3 presents the experimental results and the conclusion of the paper is given in Section 4.

## 2    Proposed System for Hand Gesture Recognition

In this section, our proposed method for recognition of hand gestures is discussed in details. The proposed system is summarized in Fig. 2. In the following, different steps of the method are explained.

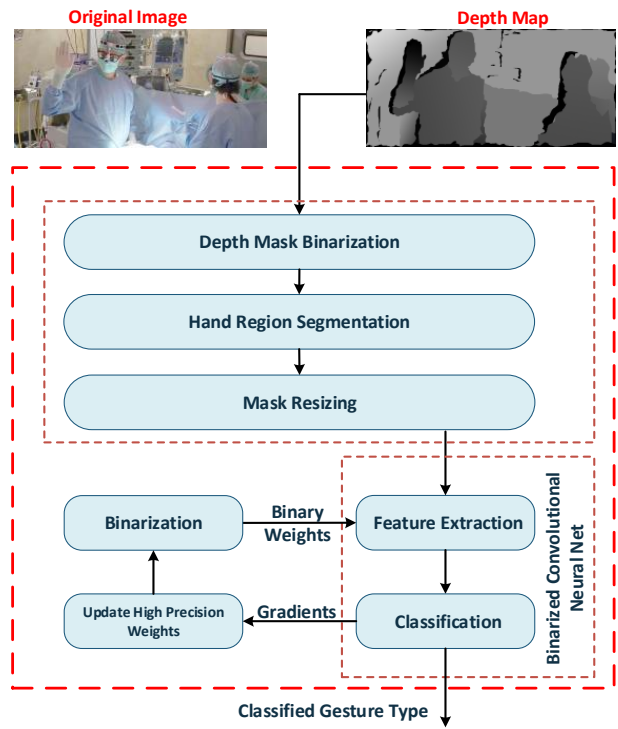

**Fig. 2.** Block diagram of our proposed training method. Example image is from [20].

## 2.1 Segmentation of Hand Region

The first stage for recognition of different hand gestures is to segment the hand region in the input image. For this aim, the depth map provided for the images in the dataset is used. The depth information is produced by the camera that takes the images. The main assumption is that hand of a person is closer to the camera as compared to the other parts of her body.

In the depth map, areas that are closer to the camera have lower values. Also it is noticed that occluded areas have zero values. Hence in order to detect the hand region, firstly the occluded regions with zero values are excluded from the image. Afterward, minimum value in the depth image, called $m$, is considered. The pixel with value $m$ would be the closest point to the camera. Since, in normal conditions, the hand of the person who is showing a gesture would be in front of his body, the point with the minimum depth value would be on the hand region. Meanwhile, usually there are small variations in the depth of different pixels on the hand area. We consider these variations as $\alpha$. Hence, pixels with the depth less than or equal to a threshold $T$, as $T = m + \alpha$, are detected as hand and are marked as 1 in a mask. The resulted mask is refined by applying dilation and hole filling morphological operations. The marked area in this binary image would include the hand region. Finally, the hand region is selected as the biggest connected component in this binary mask that includes the point with minimum depth value, $m$.

The used structure element for the dilation operation is a square with $side\ length = 3$. In our experiments, $\alpha$ is set to 3. Some sample input images, their corresponding depth maps and their produced segmentation masks are shown in Fig. 3. As can be seen, each resulted binary mask shows the gesturing hand along with the person's elbow. The obtained mask would be applied to a CNN with the architecture that is explained in the following.

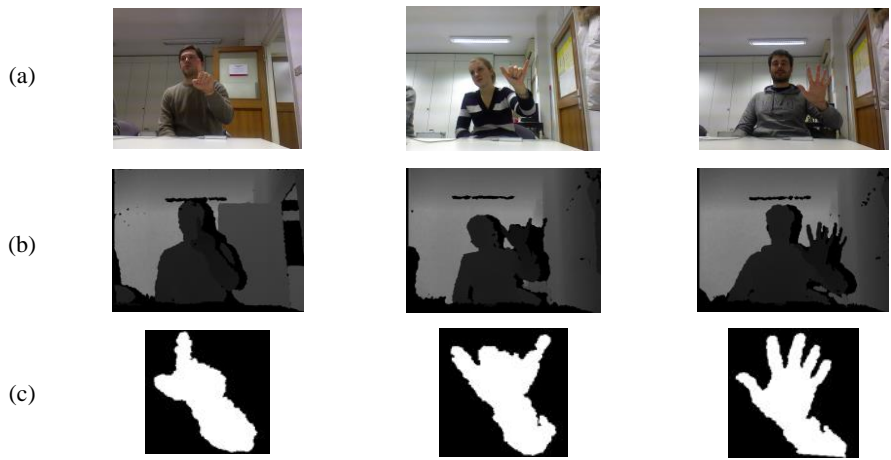

Fig. 3. Hand region segmentation for some sample images of the dataset. (a) RGB image, (b) depth map, and (c) segmented region.

## 2.2 CNN Architecture

With the aim of discriminating different hand gestures into 10 classes, the binary mask produced in the previous step is fed into a CNN with the shown architecture in Fig. 4. In order to feed an image to the network, firstly the image should be resized to 50×50. The network consists of convolution layers ($Conv1$ and $Conv2$) and max-pooling layers ($Pool1$ and $Pool2$). The order of layers in the network is as follows: $Conv1$, $Pool1$, $Conv2$, $Pool2$.

In convolutional layers, the entire input image is convolved with the same filter. This is done with the aim of extraction of same feature across the entire image. The size of each kernel in the first convolutional layer $Conv1$ is 5×5×1 and there are 50 such kernels in this layer. The second convolutional layer, i.e. $Conv2$, consists of 20 kernels with the size of 3×3×50.

In CNNs the convolutional layers are usually followed by max pooling layers. The pooling layers summarize the information that are extracted in the previous layers and facilitate the procedure of learning. The kernel sizes in the first and second pooling layers are 2×2 and 3×3, with stride of 2 and 3 respectively. This means that $Pool1$ and $Pool2$ layers result in reduction of their corresponding input size into half and 1/3.

Finally, the structure of CNN is terminated by two layers of fully connected networks. There are 50 neurons in the first and 10 neurons in the second fully connected layers. The 10 neurons in the output layer produce the membership probability of the input image to one of the 10 different existing classes of hand gestures. The neuron with highest probability indicated the type of gesture of the input image.

In the rest of this section, a method for simplification of this network for implementing it in an embedded hardware device suitable for a surgery room is presented.

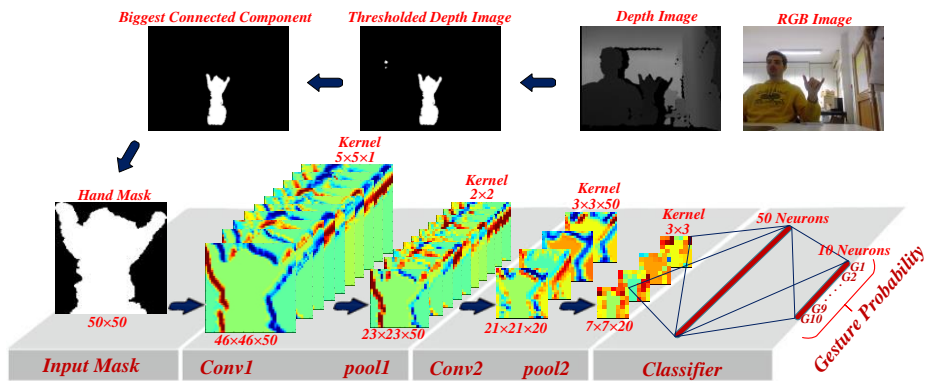

**Fig. 4.** Architecture of the employed CNN.

## 2.3 Implementation as an Embedded System

It should be noticed that the first stage of our method as segmentation of the hand region is simple enough to be implemented in a hardware device. The main bottleneck here is the complexity of the CNN as it contains millions of floating-point numbers and operations. In the following, a method for binarization of the network is used. The result would be a system that is very simple and can be embedded in any device.

Despite the high accuracy of the CNNs, one drawback in usage of these networks is that they contain millions of variables, such as weights of the kernels, which should be stored. In addition, another problem is that millions of complex floating-point operations should be done in order to calculate the output of the network. For example, there is a need for 249 MB of memory for storing the Alex-net [21] and in order to classify only one image via this net, 1.5 millions of high precision floating points operations should be performed. This requires a high performance server and cannot be implemented in a device with limited hardware capacity hence cannot be used for real-time applications.

In recent years, there has been number of researches with the aim of simplification of the CNNs, while trying to maintain the accuracy of the network [22, 23]. One approach for this aim, for reducing the network size and improving its running speed, is binarization of the network's weights in each layer of the CNN. In this paper we use a method, based on the binarization approach of [23], to facilitate the implementation of our system on an embedded hardware device. For reducing the network size, we approximate the weights in each convolutional layer as:

$$W \approx \alpha B, \qquad (1)$$

where $W$ is the matrix showing all of the weights of the kernels in a convolve layer, $\alpha$ is a floating point number that is used as a scale factor, and $B \in \{+1, -1\}^{c \times l \times h}$ is the binary approximation of the weights. Also, $c$, $l$, and $h$ are the size of the kernel's depth, width, and height, respectively.

In the following, the optimization problem in (2) should be solved to find $B$ and $\alpha$.

$$J(B, \alpha) = \| W - \alpha B \|^2 \quad s.t \ \ B \in \{+1, -1\}^{c \times l \times h}, \alpha \in \mathrm{R}, \qquad (2)$$

$$\alpha^*, B^* = \arg \min J(B, \alpha)$$

Equation (2) is solved in (3) and (4) to find optimum values for $\alpha$ and $B$ as [23]:

$$\alpha^* = \frac{1}{n} \|W\|_{l1} \qquad (3)$$

$$B^* = sing(W) \qquad (4)$$

As a result, by applying this method, the floating-point values of the network's weights would be converted to a binary matrix along with a scaling factor of $\alpha$ that is calculated for each kernel.

It should be noticed that there are two kinds of weights in our network. There are weights in the convolutional layers and the weights in the fully connected layers. For binarization of the weights of a fully connected layer, we can consider it as a convolutional layer. Hence, the number of kernels is equal to the number of its neurons and the size of the kernel would be equal to the size of the input. As a result, all of the layers in the network would be treated as convolutional layers and the weights in all of the layers would be binarized. The bias values would not be binarized.

The binarization method would be performed during the learning procedure. Since the calculated differences for weights in each run of backpropagation would be a small number. Binarizing the weights would diminish these calculated values and hinders the network's learning procedure. In order to deal with this problem, the floating point values of the weights would be stored and updated in all iterations of the learning procedure. The binarization would be performed at the beginning of all iterations. The gradients are calculated based on the binary values and these gradients. They are used to update the weights that are stored as floating-points. These floating-point values are only stored during the learning procedure. In the deployment phase, the binarized values would be stored and used. Hence these temporary stored floating-point values are not a burden for storing the network.

By binarizing the weights in all of the layers of the network, the capacity that is needed to store the network would be reduced with a scale of 32. In addition, the convolution operation would be performed based on binary values, that would result in reducing the complexity of the convolve operation. Since the weights are binarized, the multiplication operations for calculating the convolutions would be done by addition or subtraction. This would enhance the speed of the network. This achievement is gained while the accuracy of the network is decreased with a negligible value of 3 percent.

## 3    Experimental Results

To verify our work, we used a publically available dataset of [11]. This dataset consist of 10 gestures from 14 individuals. Each gesture is replicated 10 times by each individual with changes in the direction and position of the hand. Hence, initially there were 1400 images available in this dataset. In order to enhance the procedure of the learning, the initial dataset is augmented in our work. For this aim, each image in the dataset is rotated from -20 to +20 degrees with an interval of 5 degrees. As a result, 8 new synthesized images, using rotation of each image in different angles, are added to the dataset. As a result, the CNN is trained and tested using 12,600 images. In the test phase, 9 different rotations of the image (8 rotations plus the original image) are used and the final classification is done based on a voting between the 9 obtained classes.

The augmented dataset is randomly divided into 4 equal sized groups. In each run, one group is left as test and training is done based on the other 3 groups. There is no overlap between the test and the train data and the images of gestures by each person are only presented in one group. Hence, there is no common person between different groups. The CNN is implemented in Caffe [24] and Xavier method is used for initialization of weights and biases. The solver type is the Stochastic Gradient Descent (SGD).

### 3.1 CNN Training

In order to evaluate the learning procedure of the CNN, the learned kernels along with some sample feature maps of the first convolutional layer are presented in Fig. 5. The network converges to learn some features from the gesture oriented input masks of the hand. For example, it can be seen that borders of the hand in different directions are highlighted in the feature maps. It also can be inferred from the learned kernels that the network has learned some features that are related to edges of the hand and fingers. Hence it can be said that the CNN would classify the images based on the structure of the hand along the fingers extracted by defining their borders.

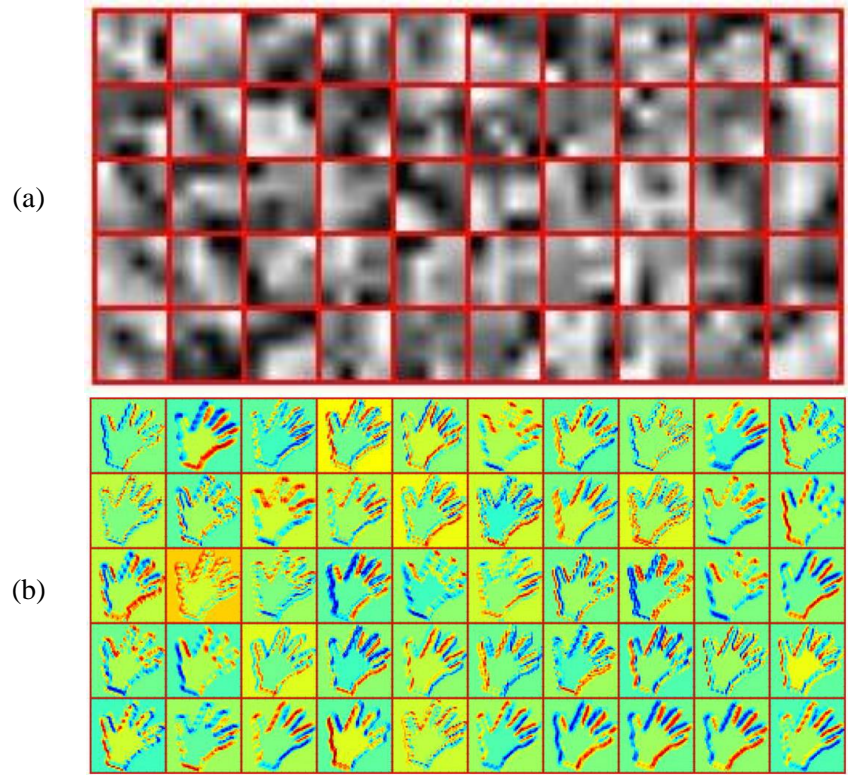

**Fig. 5.** CNN training. (a) Learned filters in the first convolutional layer and (b) sample feature maps.

### 2.1 Recognition Results

In order to evaluate our proposed system, the quantitative results of the classification done by our method are presented in Table 1. First of all, the software implementation of our method is evaluated. This is a version that the CNN is used based on its original floating-point values where the weights are not quantized. As can be seen, this version of our system produces better results in comparison to [11]. The method of [11] is a relatively complex hand gesture recognition algorithm that is not simple enough to be implemented on an embedded system.

**Table 1.** Quantitative comparison of hand gesture recognition results for 10 different gestures. Upper numbers are software results of our method and bottom numbers are from [11].

| Input \ Identified | G1 | G2 | G3 | G4 | G5 | G6 | G7 | G8 | G9 | G10 |
|---|---|---|---|---|---|---|---|---|---|---|
| G1 | **99** / 99 | 1 |   | 1 |   |   |   |   |   |   |
| G2 | 3 | **95** / 96 | 3 | 1 |   | 1 | 1 | 1 |   |   |
| G3 | 1 | 1 / 2 | **97** / 96 |   |   | 1 | 1 |   |   |   |
| G4 |   | 3 / 1 | 1 / 1 | **94** / 91 | 1 / 1 |   | 1 / 1 | 3 |   | 1 |
| G5 |   | 1 / 3 |   | 1 | **98** / 94 | 1 / 1 |   | 1 |   |   |
| G6 | 1 | 1 / 1 | 1 / 1 |   | 2 / 2 | **93** / 86 |   |   | 4 | 2 / 6 |
| G7 | 2 | 1 | 1 | 1 / 1 | 2 | 1 | **92** / 90 | 2 / 5 | 1 |   |
| G8 |   |   |   | 3 | 1 |   | 4 / 7 | **92** / 86 | 1 | 2 / 4 |
| G9 | 1 |   |   |   | 1 | 2 / 1 |   | 1 | **96** / 97 | 1 |
| G10 | 1 |   |   |   | 1 / 1 | 4 / 19 |   | 2 |   | **94** / 78 |

Moreover, the simplified version of our method, where the CNN is binarized, is also evaluated. By performing the binarization, the accuracy of our method decreases by a value of 3 percent. However, the simplified method is more desirable since its size is reduced and it could operate faster. Hence, this version would be suitable for implementation in hardware as an embedded system in a surgery room. In spite of the simplification, it can be seen in Table 2 that our method still outperforms the other method in terms of accuracy of classification for most of the gestures.

It can be seen in Table 3 that the average accuracy of classification for different gestures by our method is better than the other method, both in the complicate and simplified versions. Furthermore, it should be noticed that the variance of accuracy for different classes for our method is lower. Hence our method results in a more robust classification for different existing classes of gestures.

**Table 2.** Quantitative comparison of hand gesture recognition results for 10 different gestures. Upper numbers are from binarized implementation of our method and bottom numbers from [11].

|     | G1 | G2 | G3 | G4 | G5 | G6 | G7 | G8 | G9 | G10 |
|-----|----|----|----|----|----|----|----|----|----|----|
| G1  | **97** / **99** | 2 | 1 / 1 |   |   |   |   |   |   |   |
| G2  | 1 | **95** / **96** |  3 | 1 | 2 / 1 | 1 |   |   |   | 1 |
| G3  |   | 1 / 2 | **97** / 96 |   | 1 |   | 1 / 1 |   |   | 1 |
| G4  |   | 7 / 1 | 1 | **91** / **91** | 1 |   | 1 / 1 | 1 / 3 |   | 1 |
| G5  |   | 4 / 3 |   | 1 / 1 | **90** / **94** | 2 / 1 |   | 2 / 1 |   | 1 |
| G6  |   | 1 / 1 | 1 |   | 2 / 2 | **91** / 86 |   |   | 4 | 7 / 6 |
| G7  |   | 1 |   | 5 / 1 | 2 / 2 | 1 | 1 | **91** / 90 | 1 / 5 |   |   |
| G8  |   |   |   | 3 / 3 | 1 |   | 6 / 7 | **86** / **86** | 2 | 2 / 4 |
| G9  |   |   |   |   | 2 | 2 / 1 |   | 2 / 1 | **96** / **97** | 1 |
| G10 |   |   |   |   | 1 / 1 | 5 / 19 |   | 4 / 2 |   | **90** / 78 |

Finally, in order to complete our analysis, some sample images that are wrongly classified by our method are presented in Fig. 6. In some of these images, the gesture shown by the person lacks accuracy, i.e. the person does not properly show the gesture. In addition, in some images, due to position of the hand, the segmentation mask failed to correctly extract the hand structure. The inaccurate segmentation mask could also mislead the CNN's classification.

**Table 3.** Comparison of overall performance of gesture recognition methods.

| Method | Mean accuracy | Minimum accuracy | Variance |
|---|---|---|---|
| [11] | 91.28 | 78 | 4.2 |
| Proposed Software Method | **94.86** | **92** | **0.65** |
| Proposed Binarized Method | 92.07 | 86 | 1.4 |

## 4    Conclusion

There is a great advantage if a surgeon could control computer displays in an operating room in a touchless manner just with gestures. This could effectively reduce the risk of infection. In this paper we showed that accurate hand gesture could be achieved by deep learning. The bottleneck of implementation of deep learning in devices is the high complexity of this method. We proposed a simple and accurate implementation of deep learning is suitable for embedding purposes. We tested our method on a publically available data set and compared our results with an existing method. We proved that while our method is simple it has high accuracy.

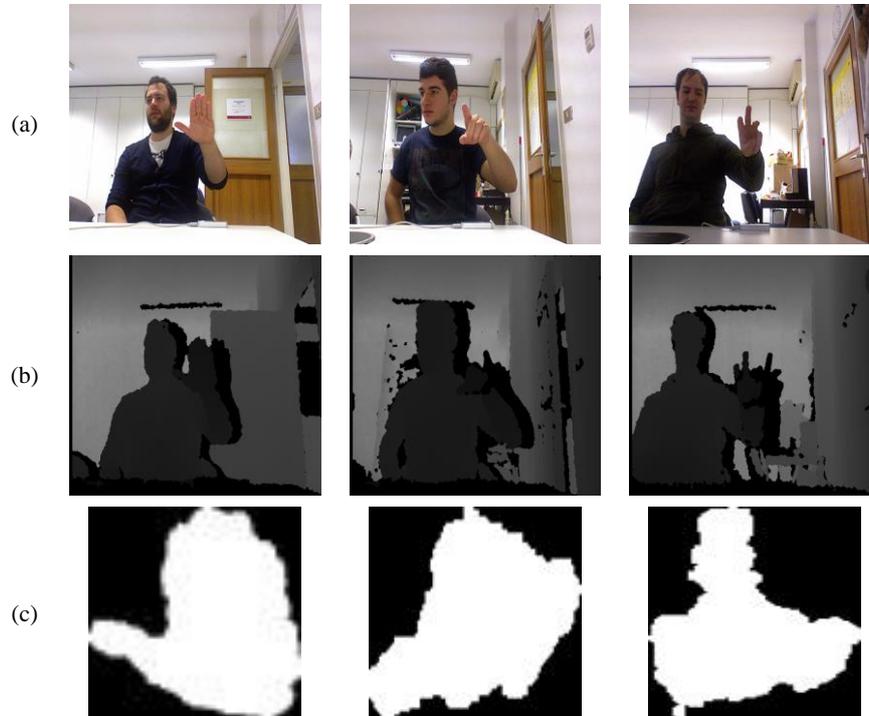

**Fig. 6.** Samples of wrongly classified gestures. (a) Input image, (b) depth map, and (c) gesture mask.